\documentclass[10pt,twocolumn,letterpaper]{article}

\usepackage[pagenumbers]{wacv} 
\usepackage[accsupp]{axessibility}

\usepackage{fancyhdr, graphics, graphicx}

\definecolor{wacvblue}{rgb}{0.21,0.49,0.74}
\usepackage[pagebackref,breaklinks,colorlinks,allcolors=wacvblue]{hyperref}
\usepackage{listings}
\usepackage[most]{tcolorbox}
\usepackage{multirow}
\usepackage{makecell}

\def\wacvPaperID{1396} 
\def\confName{WACV}
\def\confYear{2026}

\title{Seeing is Believing (and Predicting): Context-Aware Multi-Human\\Behavior Prediction with Vision Language Models}

\author{Utsav Panchal$^{1}$\thanks{Equal contribution. Work done while at Bosch Research.}\qquad
Yuchen Liu$^{1 *}$\thanks{Corresponding author: {\tt yuchen.liu@iaas.uni-stuttgart.de}}\qquad
Luigi Palmieri$^{2}$\qquad
Ilche Georgievski$^{1}$\qquad
Marco Aiello$^{1}$\\ 
\small$^{1}$ Institute of Architecture of Application Systems, University of Stuttgart, Germany\qquad$^{2}$ Bosch Research, Germany
}

\begin{document}
\maketitle

\begingroup
\renewcommand\thefootnote{}%
\footnotetext{This work was partly supported by the EU Horizon 2020 research and innovation program under grant agreement No. 101017274 (DARKO).}%
\footnotetext{Project webpage: \url{https://camp-vlm.github.io/}}
\addtocounter{footnote}{-1}%
\endgroup

\begin{abstract}
Accurately predicting human behaviors is crucial for mobile robots operating in human-populated environments. While prior research primarily focuses on predicting actions in single-human scenarios from an egocentric view, several robotic applications require understanding multiple human behaviors from a third-person perspective. 
To this end, we present CAMP-VLM (\textbf{C}ontext-\textbf{A}ware \textbf{M}ulti-human behavior \textbf{P}rediction): a Vision Language Model (VLM)-based framework that incorporates contextual features from visual input and spatial awareness from scene graphs to enhance prediction of humans-scene interactions. 
Due to the lack of suitable datasets for multi-human behavior prediction from an observer view, we perform fine-tuning of CAMP-VLM with synthetic human behavior data generated by a photorealistic simulator, and evaluate the resulting models on both synthetic and real-world sequences to assess their generalization capabilities.
Leveraging Supervised Fine-Tuning (SFT) and Direct Preference Optimization (DPO), CAMP-VLM outperforms the best-performing baseline by up to 66.9\% in prediction accuracy.
\end{abstract}
    
\section{Introduction}
\label{sec:intro}

Human behavior prediction is a critical and interdisciplinary challenge at the intersection of psychology, sociology, and computer vision \cite{rudenko2020human}, with applications spanning in autonomous driving~\cite{zhang2023pedestrian}, healthcare~\cite{phan2016deep}, smart buildings \cite{georgievski2019activity}, and especially collaborative robots~\cite{stefanini2024ral}, where accurately forecasting human behavior enables intelligent systems to make proactive decisions by anticipating high-level human actions and inferring low-level motion patterns, enhancing safe and efficient interactions, and improving coordination and navigation in dynamic environments~\cite{liu2023human}.

Over the past decade, research in human behavior prediction has progressed rapidly due to the advances in the field of Machine Learning (ML) and the availability of large-scale video datasets~\cite{kong2022human, zhong2023survey, lai2024human}. Early methods relied on Convolutional Neural Networks (CNNs)~\cite{simonyan2014two, tran2015learning} and recurrent architectures like LSTMs or RNNs~\cite{donahue2015long, abu2018will, sun2019relational, roy2022action}.
The recent emergence of Large- and Vision Language Models (LLMs/VLMs)~\cite{aiello2025paradigm, openai2023gpt4, touvron2023llama, wang2024qwen2} represents a paradigm shift. These models offer a unified framework to process information from multiple data sources and exhibit impressive generalization capabilities, making them particularly well-suited for the complexities of real-world human-robot interaction~\cite{liu2025context, zhao2024antgpt, liu2024st, chen2024motionllm}. 

\begin{figure}[t]
    \centering
    \includegraphics[width=1\linewidth]{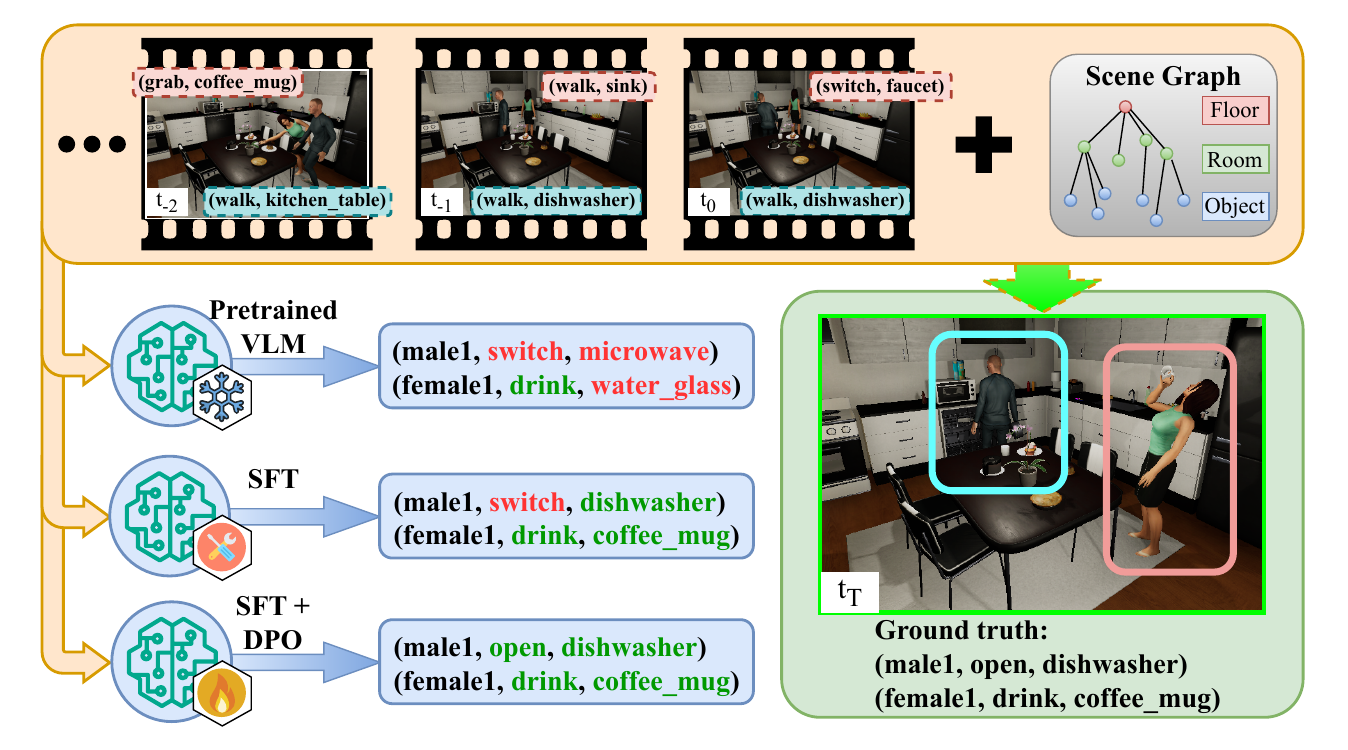}
    \caption{CAMP-VLM is a VLM-based framework for \textbf{C}ontext-\textbf{A}ware \textbf{M}ulti-human behavior \textbf{P}rediction. Receiving an image sequence of past observations from third-person views and a Scene Graph (SG) representing the environmental topologies (excluding the textual action labels), the two-stage fine-tuning process helps CAMP-VLM to more accurately predict multi-human behaviors.\vspace{-1em}}
    \label{fig:teaser}
\end{figure}

However, efforts explicitly addressing the spatial dynamics of human-scene interactions remain comparatively scarce, where spatial understanding is crucial for interpreting human-object interactions in complex environments. In contrast, Scene Graphs (SGs) offer a rich representational and hierarchical framework that encodes both semantic and spatial relationships among objects in the environment~\cite{armeni20193d}. Recent applications further demonstrate the utility of SGs in long-term human interaction prediction~\cite{gorlo2024long}, and downstream tasks such as task planning~\cite{liu2024delta, rana2023sayplan}.

While current research primarily focuses on egocentric, single-human behavior prediction~\cite{roy2022action, zhao2024antgpt, kim2024palm, beedu2024efficacy}, practical robotic applications, such as autonomous navigation and human-robot collaborations, often rely on exocentric sensors to perceive the environment, observing multiple people simultaneously, and avoiding ego-motion artifacts, as shown in Fig.~\ref{fig:teaser}. 
This setting introduces additional challenges: capturing inter-person dependencies (e.g., yielding or turn-taking), reasoning about long-range identity consistency and human–object interactions across the scene, and handling occlusion. They increase prediction uncertainty but are critical for safe and socially compliant robot operation.
Moreover, while many widely used datasets solely record single-human activities in a first-person view~\cite{kuehne2014language, grauman2022ego4d, damen2018scaling}, the availability of suitable datasets, i.e., involving scenarios with multiple humans observed from a third-person perspective, remains significantly limited.

In this work we present an attempt to bridge these critical gaps, offering the following contributions:

\begin{itemize}
    \item We propose CAMP-VLM (\textbf{C}ontext-\textbf{A}ware \textbf{M}ulti-human behavior \textbf{P}rediction), a VLM-based framework that, for the first time, integrates visual inputs with Scene Graphs for context-aware, multi-human behavior prediction from a third-person perspective.
    \item To enhance prediction accuracy and effectively capture human interactions in a data-efficient manner, we introduce a two-stage fine-tuning process combining Supervised Fine-Tuning (SFT) and Direct Preference Optimization (DPO)
    \item We provide a thorough evaluation of our method on both synthetic and real-world datasets, demonstrating a substantial improvement over existing state-of-the-art approaches. Our evaluations demonstrate that the fine-tuned model achieves up to a $66.9\%$ improvement over the best-performing baseline, signifying a substantial advancement in prediction accuracy.
\end{itemize}

CAMP-VLM represents a significant step forward in developing more socially aware and efficient robotic systems.

\section{Related Work}
\label{sec:sota}

\subsection{Human Behavior Prediction}
Behavior prediction  (or action anticipation) is the study of future human behavior, characterized as textual human-object interactions represented by \textit{verb-noun} tuples, where the noun refers to the instance of the next active object (e.g., \textit{drawer} or \textit{book}), and the verb refers to the action describing how the object instance will be used (e.g., \textit{open} or \textit{grab})~\cite{rudenko2020human}. The task of human behavior prediction focuses on capturing the correct sequential order of such tuples~\cite{kong2022human}. 
It can be categorized into egocentric and non-egocentric (i.e., from observer's perspective) predictions. The former utilizes first-person visual input and captures the scene from the viewpoint of the acting individual. The latter relies on observing the world from an external viewpoint.

\subsubsection{Prediction from Egocentric Perspective}
Egocentric view provides direct visual and contextual cues about the person's intent, hand movements, and interactions with objects~\cite{grauman2022ego4d}.
Early Deep Learning (DL)-based methods used goal-based LSTM models to predict sequential actions~\cite{roy2022action}.
Recent advances leverage LLMs for Long-Term Action Anticipation (LTA). For example, AntGPT~\cite{zhao2024antgpt} formulates LTA as a sequence modeling task using LLMs to capture intermediate and long-term goals. PALM~\cite{kim2024palm} further combines vision-language captioning and LLM-based prompting to improve predictions.
Nonetheless, egocentric perspectives of human behaviors are not suitable for several mobile robotic applications, such as navigational tasks.

\subsubsection{Prediction from Observer's Perspective}
In contrast to the egocentric view, observing the world from an observer's viewpoint is beneficial for mobile robots or autonomous vehicles.
Similarly, early efforts in third-person human behavior prediction largely relied on DL-based models, e.g., CNNs and RNNs~\cite{abu2018will, sun2019relational, tran2015learning, simonyan2014two, donahue2015long}.

\begin{figure*}[th!]
    \centering
    \includegraphics[width=0.9\linewidth]{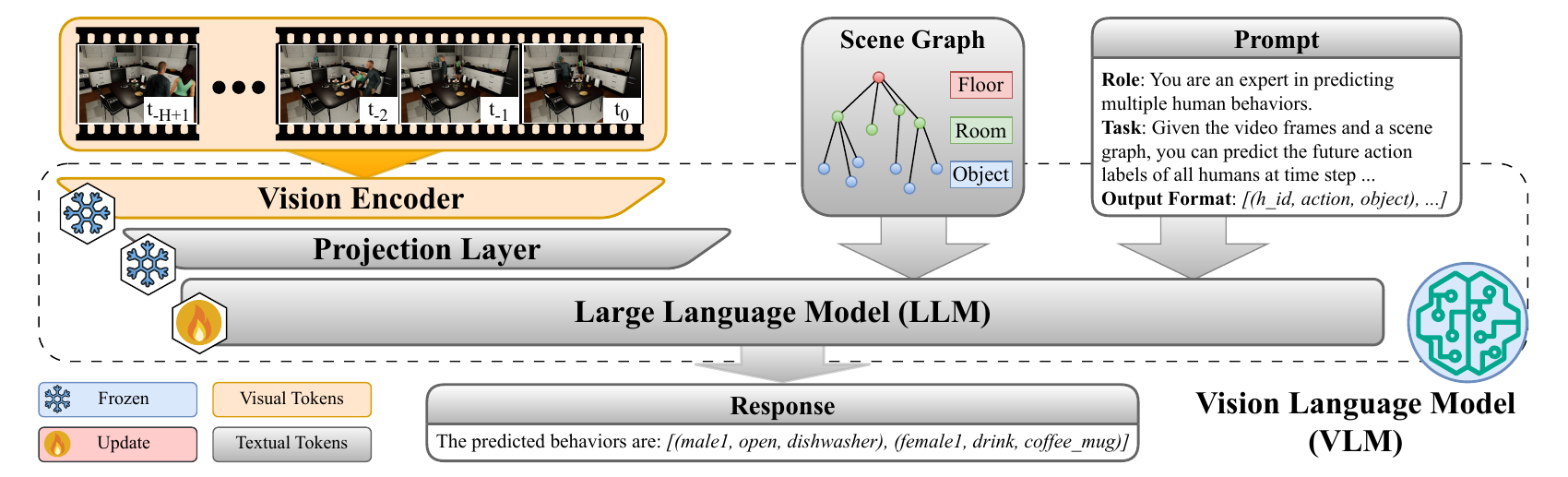}
    \caption{An overview of CAMP-VLM, a VLM-centered framework for \textbf{C}ontext-\textbf{A}ware \textbf{M}ulti-human behavior \textbf{P}rediction. The video frames are processed by the vision encoder into visual tokens, which are then passed into the Large Language Model (LLM) backbone via the projection layer. The context encoded in the images helps the VLM to discern interactions between humans and the scene. The scene knowledge encoded in the Scene Graph (SG) is provided to ground the predictions in the provided scene topologies and relationships. Under the guidance of the user-provided prompt, the LLM predicts human behaviors in the given format. The LLM is fine-tuned to improve the prediction performance, while the weights of the vision encoder and projection layer remain unchanged.\vspace{-1em}}
    \label{fig:sys}
\end{figure*}

Building on the success of LLMs and VLMs, researchers increasingly integrate them into behavior prediction systems, leveraging their zero-shot reasoning capabilities to generate semantically rich forecasts. 
Liu \etal~\cite{liu2025context} benchmarked multiple state-of-the-art pre-trained VLMs to forecast human behaviors conditioned on third-person visual scenes, historical interaction labels, and In-Context Learning (ICL) examples, demonstrating the growing potential of VLMs to unify contextual input and temporal dynamics to accurately infer human behaviors. However, the performance of pre-trained VLMs is still limited by object occlusions and challenges in retrieving object spatial relationships, resulting in a prediction accuracy lower than $70\%$.

While the majority of the existing approaches show strong progress in modeling the temporal dynamics from visual inputs (in both egocentric and third-person views), recent research highlights the growing importance of spatial awareness for more accurate and contextually grounded predictions, especially in large and context-rich environments.
Graule \etal~\cite{graule2024gg} presented GG-LLM to generate textual narrations of human activities based on visual observations. Such narrations can then be geometrically grounded in the semantic maps of the environment, thus enabling spatial awareness for downstream robot planning tasks. 
Furthermore, Gorlo \etal~\cite{gorlo2024long} demonstrated that utilizing high-level structured environment representations such as dynamic 3D scene graphs, comprising room layouts, object positions, and human affordances, leads to more effective prediction of human-object interactions over time, and reduction of uncertainty when multiple plausible actions exist based on temporal history alone. However, visual input is absent in the approach, which strongly limits the ability to recognize and temporally ground fine-grained human actions and their execution status. Combining visual perception with scene-level reasoning is essential for comprehensive behavior prediction.

\subsubsection{Multi-Human Behavior Prediction}
Despite growing interest in human behavior prediction, relatively few works focus on forecasting the behaviors of multiple humans in complex scenes. Existing methods, such as Discriminative Relational Recurrent Network (DR$^2$N)~\cite{sun2019relational}, rely on simplified representations like human bounding boxes and model interactions through pairwise relations, without incorporating environmental context or object-level semantics, and lack scene awareness.
Recent methods like HiMemFormer~\cite{wang2024himemformer} introduce memory-aware transformers to model 
multi-agent interactions, but still treat human behavior as a sequence of isolated action labels without environmental grounding. The absence of publicly available code further hinders reproducibility. These limitations highlight the need for multi-human prediction models that combine temporal reasoning with spatial context and semantic understanding.

\subsection{Datasets for Human Behavior Prediction}
Existing datasets for the aforementioned human behavior prediction methods primarily focus on egocentric perspectives and single-human scenarios, making them unsuitable for our task, i.e., predicting multi-human behaviors from a third-person view with structured scene representations. Commonly used datasets like Ego4D~\cite{grauman2022ego4d}, EPIC-Kitchens~\cite{damen2018scaling}, and EGTEA GAZE$+$~\cite{li2021eye} offer rich first-person video data but lack third-person spatial context, while third-person datasets such as Breakfast~\cite{kuehne2014language}, 50Salads~\cite{stein2013combining}, and PROX~\cite{hassan2019resolving} are limited to single-person activities. HMDB51 dataset~\cite{kuehne2011hmdb} contains a large number of movie clips with multiple actors, but is recorded in open and unconstrained spaces. A more recent dataset LEMMA~\cite{jia2020lemma} provides richly annotated multi-human activities in indoor environments from both first- and third-person views, but only features up to two agents performing concurrent tasks. Moreover, both datasets lack structured spatial environmental representations, such as scene graphs.

It is this gap that brought us to construct a new synthetic dataset using a photorealistic simulator, specifically designed to include up to three humans performing different tasks in indoor environments. To enable spatial awareness, the dataset also comprises SGs as environment representations with rich semantic relationships between objects.

\section{Methodology}
\label{sec:method}

\subsection{Task Formulation}
\label{sec:task}
The multi-human behavior prediction task involves predicting a sequence of future action labels (\textit{verb-noun} tuples) for multiple individuals within a dynamic environment. Given a sequence of past video frames $\mathbf{V}^{-H+1:0}$, with $H$ representing the frame number of observation history, and a Scene Graph (SG) $\mathcal{G}$, the task is to predict a set of future action labels $\hat{\mathbf{Y}}$ of $M$ humans up to a time horizon $T$:

\vspace{-1\baselineskip}
\begin{equation}
    \hat{\mathbf{Y}} = \left\{ \hat{y}_i^{(t)} \mid i \in M,\; t \in [1, T] \right\} = f\left( \mathbf{V}^{-H+1:0},\; \mathcal{G} \right)
\end{equation}
where $\hat{y}_i^{(t)}$ indicates the predicted action label of the $i$-th human at time step $t$.

\begin{lstlisting}[
  float=t,
  basicstyle=\fontsize{8}{8}\selectfont\ttfamily,
  caption={An example scene graph of a living room},
  label={lst:sg},
  language=Python,
  belowskip=-1\baselineskip]
living_room = {
    "tv": {
        "id": 101,
        "properties": ["HAS_SWITCH"],
        "state": ["OFF"],
        "object_placing": [{
            "destination": ("tv_stand", 102),
            "relation": "ON"
        }]
    },
    "tv_stand": {
        "id": 102,
        "properties": ["HAS_SURFACE"],
        "state": [],
        "object_placing": []
    },
    "remote_control": {
        "id": 103,
        "properties": ["GRABBABLE"],
        "state": [],
        "object_placing": [{
            "destination": ("sofa", 104),
            "relation": "ON"
        }]
    },
    "sofa": {
        "id": 104,
        "properties": ["HAS_SURFACE", "SITTABLE"],
        "state": [],
        "object_placing": []
    }
}
\end{lstlisting}

\subsection{System Architecture}
With the ambition of accurate scene- and context-aware multi-human behavior prediction, we propose CAMP-VLM illustrated in Fig.~\ref{fig:sys}, a VLM-based prediction framework that considers the following input: a sequence of video frames $\mathbf{V}^{-H+1:0}$ as visual context representations, an SG as a high-level environment representation, and a textual prompt describing the task instructions. The VLM includes three sub-components: the vision encoder processes the visual context, the projection layer maps the vision features into the text embedding space, and the language model backbone generates textual future human behavior labels $\hat{\mathbf{Y}}$. The latter will be fine-tuned, while the rest remain frozen. The fine-tuning process is explained in Sec.~\ref{sec:ft}.

\subsubsection{Visual Context Representation}
\label{sec:vis}
The video frames comprise $H$ RGB images of past observations depicting one or multiple humans from a fixed camera position, generated by a photorealistic simulation or recorded in real-world settings. 
The VLM extracts information such as the number, poses, and orientations of humans and the surrounding objects.

\subsubsection{Scene Graphs}
\label{sec:sg}
A Scene Graph (SG) is a hierarchical scene representation applied to structurally depict real-world areas. In this paper, we utilize a 2D version of SGs since most recent VLMs still use 2D inputs~\cite{wang2024qwen2}. An SG is defined as $\mathcal{G} = \{\mathcal{N}, \mathcal{E}\}$, where $\mathcal{N}$ and $\mathcal{E}$ refer to a node and an edge list, respectively, including information related to the scene. A node $n \in \mathcal{N}$ contains various attributes such as \textit{id}, \textit{properties}, and \textit{states}, where properties define a way to interact with the object, and state defines the current state of the object. An edge $e \in \mathcal{E}$ defines the spatial relationship of the object with respect to another object. 
We convert SGs into a flattened and more human-readable JSON format, where edges are represented as the \textit{object\_placing} attributes, which include the \textit{destination} and \textit{relation} sub-attributes. The former depicts the other related object, and the latter refers to the relation type. An example of the SG structure is shown in Listing~\ref{lst:sg}. A TV, a TV stand, a remote control, and a sofa are located in a living room. The TV has a property \textit{HAS\_SWITCH} and a state \textit{OFF}, indicating that it can be switched on, but it is currently turned off.

\subsubsection{Prompt Design}
\label{sec:prompt}
To provide a concise description of the general instruction to guide the behavior prediction, we specify the task context and the expected output format in the following structure:

\begin{tcolorbox}[breakable, colback=black!5!white, colframe=black!75!black, left=2pt, right=2pt, top=2pt, bottom=2pt]
\textbf{Role}: You are an expert in predicting human behaviors.

\textbf{Task}: Given $\langle H \rangle$ video frames showing multiple humans performing different actions, and a scene graph describing the available objects and their relationships in the environment, your task is to predict $\langle T \rangle$ future action labels for each human in the scene.

\textbf{Video Frames}: $\langle frame\_1.png,\; \dots,\; frame\_H.png \rangle$

\textbf{Scene Graph}: $\langle scene\_graph.json \rangle$

\textbf{Output Format}: The output should be formulated as a 2D list, where the first dimension (rows) indicates the number of humans, the second dimension (elements in each row) refers to the prediction horizon $T$, i.e., the number of future behavior labels. Each behavior label is defined as a tuple $(h\_id, \; action, \; object)$, where \textit{h\_id} means the id of the human.
\end{tcolorbox}

\subsection{Fine-Tuning Process and Dataset Preparation}
\label{sec:ft}
We select the Qwen2-VL family~\cite{wang2024qwen2}, one of the leading open-source VLMs in video understanding according to the state-of-the-art benchmarks (date 2024-11-01)~\cite{duan2024vlmevalkit, fu2025video}. To fulfill our goals, we introduce a two-stage fine-tuning process combining Supervised Fine-Tuning (SFT) and preference alignment to improve the prediction performance of pre-trained VLMs, as illustrated in Fig.~\ref{fig:ft}.

\subsubsection{Stage 1: Supervised Fine-Tuning}
\label{sec:sft}
SFT is a widely used and effective technique to adapt broadly trained LLMs or VLMs to excel at targeted tasks or styles, e.g., human behavior prediction, in a supervised learning manner. SFT allows a data-efficient learning and yields significant gains without heavy computation~\cite{pareja2024unveiling}. The low demand for cost and resources further makes it suitable for tasks with limited availability of data~\cite{bhatt2024experimental}.

A dataset $\mathcal{D}_{F} = \{(\mathbf{x}_i, \mathbf{y}_i^{gt})\}_{i=1}^K$ consisting of $K$ ground truth input-output pairs $(\mathbf{x}_i, \mathbf{y}_i^{gt})$ is prepared for the SFT process, as shown in the top-left part of Fig.~\ref{fig:ft}. 
Each input $\mathbf{x}_i$ comprises a tuple $(\mathbf{V}^{-H+1:0}, \mathcal{G}, P)$, where $P$ stands for the prediction prompt described above. The output $\mathbf{y}_i^{gt}$ refers to the ground truth behavior labels.

In this work, we employ Low-Rank Adaptation (LoRA)~\cite{hu2022lora} to the open-source pre-trained VLMs. Instead of updating all model weights, LoRA only trains the low-rank matrices to reduce the computational cost. The goal is to adapt the pre-trained base model $f_{\theta_0}$ to match the distribution of the SFT dataset $\mathcal{D}_{F}$ using the negative log-likelihood loss function as follows:

\vspace{-1\baselineskip}
\begin{equation}
    \mathcal{L}_{F}(\phi) = - \sum_{i=1}^K \log P_\phi(\mathbf{y}_i^{gt} \mid \mathbf{x}_i)
\end{equation}
where $\phi = \theta_0 + \Delta\theta$ includes the frozen base parameters $\theta_0$ and the trainable low-rank components $\Delta\theta$.

\subsubsection{Stage 2: Direct Preference Optimization}
\label{sec:dpo}
In the second stage, we further align the fine-tuned model using DPO. 
While SFT provides strong task grounding, its gains are limited by dataset size \cite{pareja2024unveiling}. Combining SFT with DPO enables the avoidance of overfitting in fine-tuned models due to small datasets, while biasing the VLM toward predictions with smaller character-level deviations.
This combination enhances the model performance by first learning core task behaviors and then optimizing outputs with human preferences, thereby further improving the response quality and task accuracy without enlarging the dataset~\cite{kirk2023understanding}, 
making it more suitable for behavior prediction tasks.

\begin{figure}[!t]
    \centering
    \includegraphics[width=1\linewidth]{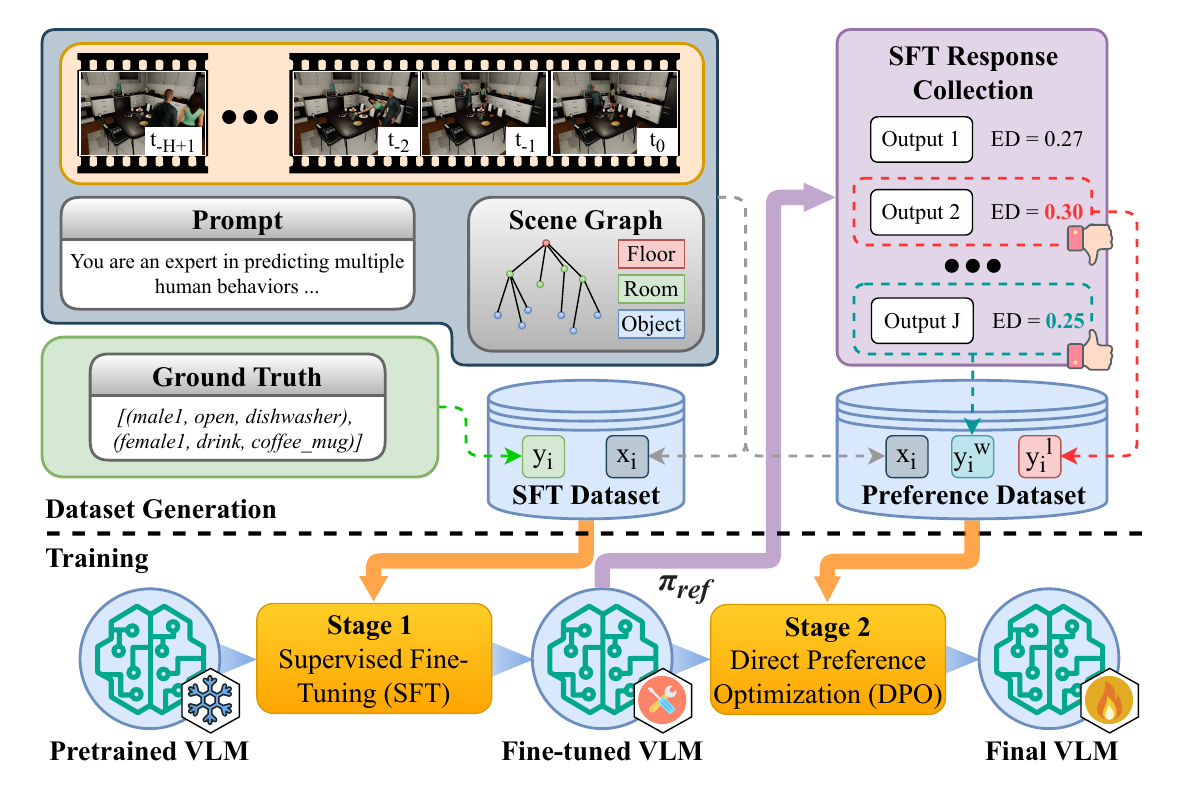}
    \caption{Data generation and fine-tuning process.\vspace{-1em}}
    \label{fig:ft}
\end{figure}

For DPO, a preference dataset $\mathcal{D}_{P} = \{(\mathbf{x}_i, \mathbf{y}_i^{w}, \mathbf{y}_i^{l})\}_{i=1}^K$ is necessary. While the input $\mathbf{x}_i$ is identical to those in $\mathcal{D}_{F}$, DPO requires additional chosen and rejected predicted labels, $\mathbf{y}_i^{w}$ and $\mathbf{y}_i^{l}$, where $w$ and $l$ refer to \textit{win} and \textit{lose}, respectively. As shown in the top-right part of Fig.~\ref{fig:ft}, to collect preferred and non-preferred data, the fine-tuned reference model $f_\text{ref}$ obtained from the last stage is inferred $J$ times to generate a batch of outputs and construct an SFT response collection. Using a pre-defined similarity metric such as Edit Distance (ED), a preferred and a non-preferred response are selected according to the lowest and highest ED values, respectively, to construct the preference dataset. The selected labels are then manually checked with the ground truth $\mathbf{y}_i^{gt}$ for any errors. This is the only step where human intervention is required.

After obtaining $\mathcal{D}_{P}$ and loading the checkpoints from the previous stage, the model is further trained by minimizing the following loss function derived from the KL-constrained reward maximization objective used in RLHF but reformulated to directly optimize the policy~\cite{rafailov2023direct}:

\vspace{-1\baselineskip}
\begin{align}
    \mathcal{L}_{P}(\pi_\theta; \pi_{\text{ref}}) = 
    & -\mathbb{E}_{(x, y^w, y^l) \sim \mathcal{D}_{P}} \Bigg[ \log \sigma\Big( \beta \log 
    \frac{\pi_\theta(y^w \mid x)}{\pi_{\text{ref}}(y^w \mid x)} \nonumber \\
    & \qquad\qquad - \beta \log 
    \frac{\pi_\theta(y^l \mid x)}{\pi_{\text{ref}}(y^l \mid x)} 
    \Big) \Bigg]
\end{align}

where $\pi_\theta(y|x)$ is the policy of the current model $f_{\theta}$ that assigns a probability to response $y$ given input $x$, $\pi_{\text{ref}}(y|x)$ is the policy of the reference model $f_\text{ref}$, $\beta$ is a hyperparameter that controls the strength of the implicit reward and the deviation from the reference policy (similar to the KL divergence penalty in RL), $\sigma(z) = 1 / (1 + \exp(-z))$ is the sigmoid function, and the term $\log \frac{\pi_\theta(y \mid x)}{\pi_{\text{ref}}(y \mid x)}$ represents the log ratio of probabilities assigned by the current policy and the reference policy to a given response.

\section{Evaluation}
\label{sec:eval}
The evaluation to assess the performance of CAMP-VLM passes through the collection of synthetic and real-world video datasets of multiple human behaviors in third-person views and conducting experiments across varying scene types and human counts. We evaluate CAMP-VLM and state-of-the-art baselines on identical inputs and report the results with selective metrics.

\begin{figure*}[!t]
    \centering
    \includegraphics[width=1\linewidth]{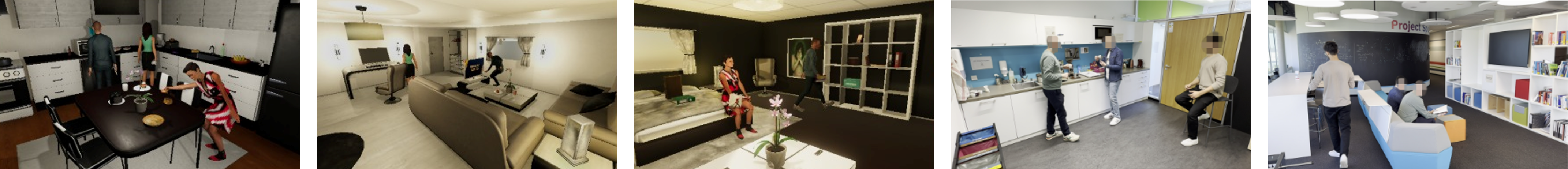}
    \caption{Example scenes of the datasets. From left to right: kitchen, living room, bedroom from VirtualHome simulation~\cite{puig2018virtualhome}, and office kitchen and living room in the real-world video recordings.\vspace{-1em}}
    \label{fig:scene}
\end{figure*}

\subsection{Data Collection}
\label{sec:data}
As the existing datasets do not meet the requirements for multi-human behavior prediction in third-person views, we generate a synthetic dataset using the VirtualHome simulator~\cite{puig2018virtualhome}, and also record videos in the real world. The former is used for the fine-tuning process, and both datasets are used for evaluation. VirtualHome is a 3D photorealistic simulation built upon the Unity game engine. It includes $7$ different household environments, and can simulate $6$ humanoid characters with $18$ actions with user-defined action scripts, which consist of a sequence of symbolic actions in the form of $\langle char\_id \rangle \; [action] \; \langle object \rangle \; (object\_id)$, where $\langle char\_id \rangle$ corresponds to the id of the character. $[action]$ and $\langle object \rangle$ can be chosen from the supported ones pre-defined in the simulator, and $(object\_id)$ further specifies the object instance in the environment to interact with.
We generate $30$ synthetic video sequences, including $20$ videos spanning $3$ different room types (i.e., kitchen, bedroom, and living room) for $1$ and $2$ humans, respectively, and $10$ videos for $3$ humans in the kitchen and living room. Each video lasts up to $30s$, and is rendered at $15$ frames-per-second (fps). The video frames are uniformly sampled as RGB images with $0.5s$ time interval based on an informal validation to balance temporal action coverage with token budget for VLMs, which is a common practice in many human behavior datasets \cite{damen2018scaling}.
The SGs can be generated by the simulator using the $environment\_graph()$ function, and converted to the desired format mentioned in Sec.~\ref{sec:sg}.

We also recorded $10$ real-world videos featuring $2$ to $3$ humans in an office kitchen and a living room, captured with an iPhone 14 Pro at $30$ fps from a fixed position, each sequence ranges up to $1$ minute duration. The total duration of both synthetic and real-world videos amounts to circa $22$ minutes.
Table~\ref{tab:data_stats} depicts the major statistics of the datasets. All actions in both datasets have an equal chance level.
Fig.~\ref{fig:scene} shows some example scenes in both datasets. 

\begin{table}[t]
\centering
\small
\renewcommand{\arraystretch}{0.9}
\setlength{\tabcolsep}{3pt}
\begin{tabular}{*{6}{c}}
\toprule
\textbf{Type} & \textbf{\#Human} & \textbf{Room} & \textbf{\#Video} & \textbf{Avg. Dur.$[s]$} & \textbf{\#Action} \\
\midrule
\multirow{9}{*}{\makecell{Syn.\\($15$ fps)}} 
  & \multirow{3}{*}{1} & K     & 4 & 30 & 6 \\
  &                    & B     & 3 & 23 & 4 \\
  &                    & L     & 3 & 20 & 4 \\
\cmidrule(lr){2-6}
  & \multirow{3}{*}{2} & K     & 4 & 30 & 13 \\
  &                    & B     & 3 & 23 & 9 \\
  &                    & L     & 3 & 20 & 9 \\
\cmidrule(lr){2-6}
  & \multirow{2}{*}{3} & K     & 5 & 25 & 13 \\
  &                    & L     & 5 & 23 & 9 \\
\midrule
\multirow{4}{*}{\makecell{Real\\World\\($30$ fps)}} 
  & \multirow{2}{*}{2} & K     & 4 & 60 & 9 \\
  &                    & L     & 2 & 52 & 8 \\
\cmidrule(lr){2-6}
  & \multirow{2}{*}{3} & K     & 2 & 50 & 9 \\
  &                    & L     & 2 & 60 & 8 \\
\bottomrule
\end{tabular}
\caption{Statistics of the datasets. The rooms cover K(itchen), B(bedroom), and L(iving room). \textbf{Duration} and the number of \textbf{Actions} per human are averaged through all videos.\vspace{-1em}}
\label{tab:data_stats}
\end{table}

\subsection{Model Selection and Implementation}
To examine the impact of different model sizes on the prediction performance, we explore three different variants: Qwen2-VL-72B-, -7B-, and -2B-Instruct (in the following denoted as Qwen-72B, -7B, and 2B). We evaluate CAMP-VLM mainly with the 72B variant, while the rest are used for the ablation study. For the SFT process, the models are fine-tuned with $30$ epochs using the AdamW~\cite{loshchilov2017decoupled} optimizer with \textit{adamw\_torch\_fused} configuration. Constant learning rate scheduler is applied with learning rate set to $3e-4$. As explained in Sec.~\ref{sec:ft}, LoRA~\cite{hu2022lora} is used in the SFT stage with rank ($r$) and alpha ($\alpha$) set to $64$ and $128$, correspondingly. For the DPO stage, the models are trained with $5$ epochs using the same optimizer, as the input data are already seen in the last stage to avoid overfitting. In both stages, the temperature is set to $1.0$ and bfloat16 precision is used to allow more diverse responses while reducing the memory demand. The Qwen models are trained on three and deployed on two Nvidia H100 GPUs.

We adopt a $70 / 30$ train-test dataset split. Both the number of input frames and the prediction horizon are set to $H = T = 6$ (i.e., $3s$) for all experiments.

\subsection{Baselines}
We compare CAMP-VLM to two representative baselines.
\textbf{AntGPT} is a state-of-the-art method for action anticipation, leading the Ego4D LTA benchmark~\cite{ego4d_ltaa}. 
Despite being trained on egocentric data, AntGPT is a transferable architecture for action anticipation, which remains applicable to exocentric, multi-human forecasting. It also significantly outperforms other DL- and transformer-based approaches.
Including AntGPT allows an informative comparison by testing whether spatial and third-person temporal context can add value over a general-purpose LLM-based prediction pipeline.
We fine-tuned AntGPT with our synthetic dataset.

We also select the best-performing configuration in a recent benchmark on context-aware human behavior prediction (in the following denoted as \textbf{CAP}) as the second baseline~\cite{liu2025context}. CAP is evaluated on the PROX dataset~\cite{hassan2019resolving}, a collection of single-human behavior videos from observer perspectives. CAP reveals that combining GPT-4o with autoregressive prediction and maximal-allowed ICL examples leads to the highest prediction accuracy. Since each input video sequence consists of $6$ images, and GPT-4o allows a maximum $50$ images in a prompt, together with a query, $7$ is the maximum possible number of examples ($(7 + 1) \times 6 = 48 < 50$). 
GPT-4o (version 2024-05-13) is deployed using Azure OpenAI Service.

\subsection{Metrics}
To support an open-vocabulary prediction paradigm, we evaluate the models using the following metrics: \textbf{Accuracy score} measures the degree of overlap between the predicted and ground truth behavior labels, ranging from $0$ to $1$, where $1$ indicates the exact match. We also evaluate partial accuracy with respect to verbs and nouns. \textbf{Cosine similarity} reports the semantic similarity of two strings by computing the cosine of the angle between their vector embeddings. The value lies between $-1$ and $1$, where $1$ denotes identical meaning. \textbf{Edit distance} examines the character-level similarity between the prediction and the ground truth by calculating the normalized minimum number of operations (deletions, insertions, or substitutions) to transform one string into another. The value varies from $0$ to $1$, where the lower value indicates a better match.

\section{Results and Discussion}
\label{sec:result}

\begin{table}[t]
\centering
\small
\renewcommand{\arraystretch}{0.9}
\setlength{\tabcolsep}{2.5pt}
\begin{tabular}{*{8}{c}}
\toprule
\multirow{2}{*}{\makecell{\textbf{\#Hu-}\\\textbf{man}}} & \multirow{2}{*}{\makecell{\textbf{Room}\\(Syn.)}} & \multirow{2}{*}{\textbf{Model}} & \multicolumn{3}{c}{\textbf{Accuracy $\uparrow$}} & \multirow{2}{*}{\textbf{CS\textbf{ $\uparrow$}}} & \multirow{2}{*}{\textbf{ED $\downarrow$}} \\
\cmidrule(lr){4-6}
& & & Full & Verb & Noun & & \\
\midrule
\multirow{9}{*}{2}  & \multirow{3}{*}{K} & AntGPT   & 0.053 & 0.108 & 0.246 & 0.961 & 0.503 \\
                    &                    & CAP      & 0.205 & 0.466 & 0.307 & 0.963 & 0.351 \\
                    &                    & CAMP-VLM & \textbf{0.482} & \textbf{0.587} & \textbf{0.610} & \textbf{0.967} & \textbf{0.262} \\
\cmidrule(lr){2-8}
                    & \multirow{3}{*}{L} & AntGPT   & 0.079 & 0.342 & 0.173 & 0.953 & 0.349 \\
                    &                    & CAP      & 0.183 & 0.368 & 0.203 & 0.956 & 0.347 \\
                    &                    & CAMP-VLM & \textbf{0.276} & \textbf{0.468} & \textbf{0.384} & \textbf{0.965} & \textbf{0.259} \\
\cmidrule(lr){2-8}
                    & \multirow{3}{*}{B} & AntGPT   & 0.128 & 0.403 & 0.309 & \textbf{0.969} & 0.291 \\
                    &                    & CAP      & 0.213 & 0.453 & 0.318 & 0.967 & 0.297 \\
                    &                    & CAMP-VLM & \textbf{0.398} & \textbf{0.568} & \textbf{0.513} & 0.965 & \textbf{0.203} \\
\midrule
\multirow{6}{*}{3} & \multirow{3}{*}{K} & AntGPT   & 0.093 & 0.115 & 0.154 & 0.952 & 0.408 \\
                    &                   & CAP      & 0.137 & 0.309 & 0.320 & 0.962 & 0.374 \\
                    &                   & CAMP-VLM & \textbf{0.301} & \textbf{0.473} & \textbf{0.439} & \textbf{0.963} & \textbf{0.328} \\
\cmidrule(lr){2-8}
                    & \multirow{3}{*}{L} & AntGPT   & 0.103 & 0.152 & 0.137 & 0.947 & 0.451 \\
                    &                    & CAP      & 0.172 & 0.286 & 0.301 & 0.961 & 0.324 \\
                    &                    & CAMP-VLM & \textbf{0.227} & \textbf{0.395} & \textbf{0.316} & \textbf{0.963} & \textbf{0.305} \\
\bottomrule
\end{tabular}
\caption{Results of all models in multi-human scenarios across all \textbf{synthetic} K(itchen), B(bedroom), and L(iving room) scenes. Higher values of accuracy and Cosine Similarity (CS) and lower values of Edit Distance (ED) indicate better performance. Bold numbers are the best results in each sub-category.\vspace{-1em}}
\label{tab:result_syn}
\end{table}

\begin{table}[t]
\centering
\small
\renewcommand{\arraystretch}{0.9}
\setlength{\tabcolsep}{2.5pt}
\begin{tabular}{*{8}{c}}
\toprule
\multirow{2}{*}{\makecell{\textbf{\#Hu-}\\\textbf{man}}} & \multirow{2}{*}{\makecell{\textbf{Room}\\(Real.)}} & \multirow{2}{*}{\textbf{Model}} & \multicolumn{3}{c}{\textbf{Accuracy $\uparrow$}} & \multirow{2}{*}{\textbf{CS\textbf{ $\uparrow$}}} & \multirow{2}{*}{\textbf{ED $\downarrow$}} \\
\cmidrule(lr){4-6}
& & & Full & Verb & Noun & & \\
\midrule
\multirow{4}{*}{2}  & \multirow{2}{*}{K} & CAP      & 0.326 & 0.357 & 0.385 & 0.953 & 0.386 \\
                    &                    & CAMP-VLM & \textbf{0.425} & \textbf{0.486} & \textbf{0.421} & \textbf{0.965} & \textbf{0.310} \\
\cmidrule(lr){2-8}
                    & \multirow{2}{*}{L} & CAP      & 0.240 & 0.300 & 0.338 & 0.960 & 0.386 \\
                    &                    & CAMP-VLM & \textbf{0.362} & \textbf{0.407} & \textbf{0.425} & \textbf{0.963} & \textbf{0.330} \\
\midrule
\multirow{4}{*}{3}  & \multirow{2}{*}{K} & CAP      & 0.305 & 0.392 & 0.406 & 0.953 & 0.403 \\
                    &                    & CAMP-VLM & \textbf{0.410} & \textbf{0.451} & \textbf{0.462} & \textbf{0.962} & \textbf{0.327} \\
\cmidrule(lr){2-8}
                    & \multirow{2}{*}{L} & CAP      & 0.238 & 0.285 & 0.315 & 0.952 & 0.392 \\
                    &                    & CAMP-VLM & \textbf{0.334} & \textbf{0.392} & \textbf{0.361} & \textbf{0.960} & \textbf{0.340} \\
\bottomrule
\end{tabular}
\caption{Results of CAMP-VLM and CAP in \textbf{real-world} multi-human scenarios in K(itchen) and L(iving room). Results of AntGPT are not listed due to low performance.\vspace{-1em}}
\label{tab:result_real}
\end{table}

\textbf{CAMP-VLM vs. Baselines.}
Across all tested multi-human cases shown in Tables~\ref{tab:result_syn} and \ref{tab:result_real}, CAMP-VLM consistently outperforms the baseline models by a substantial margin. 
E.g., in the synthetic kitchen scene with two humans, CAMP-VLM reaches the highest $0.482$ full accuracy.
A similar trend is also observed in other synthetic and real-world scenes with two humans, where CAMP-VLM is $66.9\%$ ($0.389$ vs. $0.233$, averaged on both synthetic and real-world results) and over $3$ times ($0.389$ vs. $0.087$) more accurate than CAP and AntGPT, respectively.
In the most complex scene, i.e., the three-human synthetic kitchen due to the highest number of humans and action per human according to Table~\ref{tab:data_stats}, the margin remains wide, where CAMP-VLM still leads the performance in all metrics, exceeding the best-performing baseline (CAP) by $49.3\%$ ($0.318$ vs. $0.213$) on average, while AntGPT still struggles with the lowest accuracies around $0.1$.
Moreover, CAMP-VLM also produces consistently lower Edit Distance, for example, down to $0.203$ in the two-human synthetic bedroom. AntGPT yields the highest cosine similarity in the same scene, but it is marginally higher than CAMP-VLM by $0.004$. 
These metrics indicate that CAMP-VLM can predict human behaviors that align more closely with the ground truth, whereas the baselines deviate more.

\begin{table}[t]
\centering
\small
\renewcommand{\arraystretch}{0.9}
\setlength{\tabcolsep}{2.6pt}
\begin{tabular}{l *{5}{c}}
\toprule
\multirow{2}{*}{\textbf{Model}} & \multicolumn{3}{c}{\textbf{Accuracy $\uparrow$}} & \multirow{2}{*}{\textbf{CS\textbf{ $\uparrow$}}} & \multirow{2}{*}{\textbf{ED $\downarrow$}} \\
\cmidrule(lr){2-4}
& Full & Verb & Noun & & \\
\midrule
Qwen-72B (Pretrained) & 0.103 & 0.156 & 0.127 & 0.874 & 0.692 \\
Qwen-72B (SFT)        & 0.294 & 0.392 & 0.328 & 0.958 & \textbf{0.315} \\
\midrule
Qwen-72B (SFT+DPO)    & \textbf{0.301} & \textbf{0.473} & \textbf{0.439} & \textbf{0.963} & 0.328 \\
\midrule
Qwen-7B (SFT+DPO)     & 0.144 & 0.206 & 0.198 & 0.942 & 0.425 \\
Qwen-2B (SFT+DPO)     & 0.127 & 0.153 & 0.174 & 0.941 & 0.437 \\
\bottomrule
\end{tabular}
\caption{Performance of CAMP-VLM with different fine-tuning strategies and VLM variants in kitchen (syn.) with 3 humans.\vspace{-1em}}
\label{tab:result_ft}
\end{table}

\textbf{Performance with Different Fine-Tuning Strategies.}
A key ablation is how the fine-tuning strategy impacts the performance. Using the most complex scene (three-person synthetic kitchen) as an example, the pretrained model barely manages $0.103$ full accuracy with low verb ($0.156$) and noun ($0.127$) precisions, as shown in the upper part of Table~\ref{tab:result_ft}. SFT yields a notable improvement, where the full accuracy increased by $185\%$ (to $0.294$), and edit distance reduced by $54.5\%$ ($0.692$ to $0.315$). This indicates that the model learns to recognize relevant patterns for the task from the synthetic training examples, and the predictions become much closer to the ground truth.
Adding DPO delivers a further performance boost. Although the full accuracy rises marginally ($0.301$ vs. $0.294$), verb and noun accuracies remarkably climb to $0.473$ and $0.439$, i.e., increased by $20.6\%$ and $33.8\%$ compared to SFT, respectively, resulting in nearly tripling the performance of the pretrained model. Cosine similarity increased slightly, suggesting the pretrained model already produces semantically plausible outputs. Overall, SFT provides a strong task grounding, and DPO further enhances output precision and alignment.

\textbf{Performance with Different VLM Variants.}
We also investigate how the size of the VLM backbone affects prediction performance based on the synthetic kitchen scene with three humans, as depicted in the middle and lower sections of Table~\ref{tab:result_ft}. Increasing the size of Qwen2-VL, the core component of CAMP-VLM, from 2B to 72B, one can observe a clear trend that larger model capacity drastically improves the performance. Even using SFT and DPO, the smallest 2B variant only achieves $0.127$ full accuracy, while the 7B variant is slightly better ($0.144$). In contrast, the 72B variant, which is used in CAMP-VLM, is $235\%$ more accurate than the 7B one. The results emphasize that model size is critical for multi-human behavior prediction, a challenging visual understanding task. The larger VLM (72B) can better capture the nuances of multi-human interactions than the smaller ones, leading to superior performance.

\begin{table}[t]
\centering
\small
\begin{tabular}{*{6}{c}}
\toprule
\multirow{2}{*}{\textbf{\#Human}} & \multicolumn{3}{c}{\textbf{Accuracy $\uparrow$}} & \multirow{2}{*}{\textbf{CS\textbf{ $\uparrow$}}} & \multirow{2}{*}{\textbf{ED $\downarrow$}} \\
\cmidrule(lr){2-4}
& Full & Verb & Noun & & \\
\midrule
1 & \textbf{0.529} & \textbf{0.588} & \textbf{0.581} & \textbf{0.969} & \textbf{0.194} \\
2 & 0.389          & 0.503          & 0.471          & 0.965          & 0.273 \\
3 & 0.318          & 0.428          & 0.395          & 0.962          & 0.325 \\
\bottomrule
\end{tabular}
\caption{Performance of CAMP-VLM with increasing number of humans. The values are averaged across all room types.\vspace{-1em}}
\label{tab:result_human}
\end{table}

\textbf{Performance with Increasing Number of Humans.}
Results in Table~\ref{tab:result_human} imply the growing difficulty in behavior prediction with an increasing number of humans, leading to performance degradation from $0.529$ (one human) to $0.389$ (two humans) and $0.318$ (three humans). Nonetheless, Table~\ref{tab:result_syn} shows that CAMP-VLM still outperforms baselines at every level in multi-human cases. 
Despite this, CAMP-VLM maintains good prediction quality, with cosine similarity above $0.96$ and a moderate increase in edit distance, e.g., from $0.194$ (one human) to $0.273$ (two humans) and $0.325$ (three humans).
The results indicate that the difficulties and challenges rise significantly in handling the combinatorial complexity of multi-human behavior predictions with the growing number of humans.

\begin{table}[t]
\centering
\small
\begin{tabular}{*{6}{c}}
\toprule
\multirow{2}{*}{\makecell{\textbf{Presence}\\\textbf{of SG}}} & \multicolumn{3}{c}{\textbf{Accuracy $\uparrow$}} & \multirow{2}{*}{\textbf{CS\textbf{ $\uparrow$}}} & \multirow{2}{*}{\textbf{ED $\downarrow$}} \\
\cmidrule(lr){2-4}
& Full & Verb & Noun & & \\
\midrule
w/o SG & 0.105          & 0.318          & 0.172          & 0.918          & 0.398 \\
w. SG  & \textbf{0.183} & \textbf{0.462} & \textbf{0.305} & \textbf{0.971} & \textbf{0.355} \\
\bottomrule
\end{tabular}
\caption{Comparison of performance with and without scene graphs using pretrained GPT-4o in kitchen (syn.) with 3 humans.\vspace{-1em}}
\label{tab:result_sg}
\end{table}

\textbf{Impact of Scene Graphs.}
To better analyze the performance gain with SGs on adding spatial awareness, we conduct a stand-alone experiment without inputting SGs. To visualize the direct impact of the SGs and eliminate the effect brought by the training data, we replace the core VLM with pretrained GPT-4o (due to the poor performance of pretrained Qwen-72B shown in Table~\ref{tab:result_ft}), and guide the reasoning with a maximum of $7$ ICL examples with SGs. Again, in the most complex scenario (synthetic kitchen with three humans), adding SG provides $45.3\%$ and $77.3\%$ improvements in verb and noun accuracies, respectively, as shown in Table~\ref{tab:result_sg}. The larger gain in noun accuracy implies notably better object grounding capability, which is contributed by including the spatial relationships between objects in SGs.

\begin{table}[t]
\centering
\small
\renewcommand{\arraystretch}{0.9}
\begin{tabular}{*{6}{c}}
\toprule
\multirow{2}{*}{\makecell{\textbf{Scene}\\\textbf{Type}}} & \multicolumn{3}{c}{\textbf{Accuracy $\uparrow$}} & \multirow{2}{*}{\textbf{CS\textbf{ $\uparrow$}}} & \multirow{2}{*}{\textbf{ED $\downarrow$}} \\
\cmidrule(lr){2-4}
& Full & Verb & Noun & & \\
\midrule
Synthetic  & 0.337          & \textbf{0.498} & \textbf{0.452} & \textbf{0.965} & \textbf{0.271} \\
Real-World & \textbf{0.383} & 0.434          & 0.417          & 0.963          & 0.328 \\
\bottomrule
\end{tabular}
\caption{Comparison of prediction performance of CAMP-VLM between synthetic and real-world scenes. The values are averaged across all room types in 2 and 3 humans scenarios.\vspace{-1em}}
\label{tab:syn_vs_real}
\end{table}

\textbf{Synthetic vs.\ in Real-World Scenes.}
As shown in Table~\ref{tab:syn_vs_real}, CAMP-VLM achieves slightly higher full accuracy in real-world scenes ($0.383$ vs. $0.337$), indicating good generalization capability. However, partial accuracies, cosine similarity, and edit distance are still better in synthetic settings, likely due to domain alignment with the training data. In real videos, the model more often mispredicts nouns ($0.452$ for synthetic vs. $0.417$ for real videos), possibly due to challenges in occlusions and object recognition.

\textbf{Limitations.}
While CAMP-VLM achieves promising results, its performance is limited by the size of the training dataset, which restricts generalization to more diverse or complex scenarios. This is particularly evident in the three-human settings, where the accuracy remains relatively low (around $0.3$), highlighting the inherent difficulty of multi-human behavior prediction from observer perspectives. Additionally, VLMs often rely on surface-level correlations and struggle to model deeper cognitive processes. Finally, our method assumes access to 2D scene graphs for spatial context, but integrating 3D environment information still remains an open challenge and potential future work.

\section{Conclusion}
\label{sec:conclusion}
Predicting multi-human behaviors from third-person views in complex environments is a challenging task. 
To confront this, we introduce CAMP-VLM: a VLM-based framework for \textbf{C}ontext-\textbf{A}ware \textbf{M}ulti-human behavior \textbf{P}rediction, integrating contextual features from visual input and spatial relationships from scene graphs, and fine-tuned with SFT and DPO. 
Evaluated across synthetic and real-world datasets, CAMP-VLM outperforms the best-performing baseline by up to $66.9\%$. 
Despite remaining challenges in dataset scale and multi-agent interaction modeling, CAMP-VLM offers a promising step toward scalable third-person multi-human behavior prediction.
{
    \small
    \bibliographystyle{ieeenat_fullname}
    \bibliography{main}
}

\end{document}